\documentclass{article}

\usepackage{hyperref}

\newcommand{\gap}{ [\dots] }

\begin{document}

\title{Demolishing Searle's Chinese Room}
\author{Wolfram Schmied\\\href{mailto:wolfram.schmied@gmx.net}{wolfram.schmied@gmx.net}}
\date{March 8, 2004}
\maketitle

\abstract{Searle's Chinese Room argument is refuted by showing that he has actually given two different versions of the room, which fail for different reasons. }

\section{Introduction}
In \cite{Searle80a}, Searle tried ``to show that a system could have input and output capabilities that duplicated those of a native Chinese speaker and still not understand Chinese''. To do so, he conducted a \emph{Gedankenexperiment} known as the Chinese Room which has been widely discussed thereafter \cite[Introduction]{Hauser93} \cite[Introduction]{Harnad89}.\footnote{A Google search for +Searle +``Chinese room'' yields 7000+ hits on the WWW, and 4000+ on Usenet.} But even after 24 years, there seems to be no general agreement on Searle's argument that computers are incapable of understanding. In the hope of putting this argument to rest, I will show that the argument actually consists of two different versions, each of which fails for a different reason. In contrast to earlier critiques, I reconstruct only where absolutely necessary, sticking as close to Searle's text as possible, concentrating solely on refuting the argument. I specifically do not take any position on whether computers can understand, have cognitive states, think, are human, have intentionality, etc. I'm assuming familiarity with Searle's paper, and have provided links to online sources for all references in the bibliography.

\section{The Chinese Room Argument}
This is the original version:
\begin{quotation}
	I do not understand a word of the Chinese stories. \emph{I have inputs and 
	outputs that are indistinguishable from those of the native Chinese speaker,} 
	\gap but I still understand nothing. For the same reasons, Schank's computer 
	understands nothing of any stories, \gap since 
	in the Chinese case the computer is me, and in cases where the computer is not me, 
	the computer has nothing more than I have in the case where I understand nothing.
\end{quotation}
The problem is with the sentence I emphasized. Not \emph{Searle}, the \emph{room} creates outputs indistinguishable from those of native speakers. To see this, just remove the program from the room. Searle himself stays unaffected, but Chinese queries will no longer be correctly answered by the room. To clarify: It is not computers that return output for input, it is computers running a program. So a computer running the Chinese program \emph{does} have something that Searle doesn't, and that is the Chinese program. Since the analogy between Searle and a computer fails because Searle is not running the same program as the computer, the conclusion that ``Schank's computer understands nothing of any stories'' is no longer supported.

Searle commits the mistake of confusing himself and the room already when he describes the room:
\begin{quotation}
	--- that is, from the point of view of somebody outside the room in which I am locked 
	--- \emph{my} answers to the questions are absolutely indistinguishable from those of 
	native Chinese speakers. Nobody just looking at \emph{my} answers can tell that I don't 
	speak a word of Chinese.
\end{quotation}

\section{The Internalized Chinese Room Argument}
This version of the room is used in the rejoinder to the systems reply:
\begin{quotation}
	Let the individual internalize all of these elements of the system. He memorizes the rules 
	in the ledger and the data banks of Chinese symbols, and he does all the calculations in 
	his head. The individual then incorporates the entire system. \gap 
	\emph{All the same, he understands nothing of the Chinese}, 
	and a fortiori neither does the system, because there isn't anything in the system that 
	isn't in him. If he doesn't understand, then there is no way the system could understand 
	because the system is just a part of him.
\end{quotation}
In this case, it is indeed Searle who produces Chinese output. But now he can no longer claim to not understand Chinese! To see this, note that the Searle of the previous section did not speak any Chinese at all, and therefore \emph{could} not understand Chinese stories, using the (uncontroversial) assumption that speaking a language is necessary for understanding stories in that language. But if Searle internalizes the Chinese program, we can no longer deny that he speaks Chinese, therefore we need an independent reason to deny that he understands Chinese. Searle does not give any such reason, again failing to support his conclusion.

\section{Summary}
Searle created two versions of the Chinese Room argument. The first version fails because Searle, as opposed to the room containing him and the Chinese program, is not equivalent to a computer running the Chinese program, therefore his failure to understand Chinese does not force the same failure upon the computer. The second version fails because Searle failed to establish a reason for denying understanding of a language of systems capable of speaking that language.

\bibliographystyle{plain}
\bibliography{ChineseRoom}

\end{document}